\documentclass{article}     

\usepackage{arxiv}
\usepackage[utf8]{inputenc} 
\usepackage[T1]{fontenc}    
\usepackage{hyperref}       
\usepackage{url}            
\usepackage{booktabs}       
\usepackage{amsfonts}       
\usepackage{nicefrac}       
\usepackage{microtype}      
\usepackage{lipsum}
\usepackage{graphicx}
\usepackage{caption}
\usepackage{subcaption}
\usepackage{multirow}
\usepackage{graphics} 
\usepackage{mathptmx} 
\usepackage{times} 
\usepackage{amsmath} 
\usepackage{amssymb}  
\usepackage{pifont}
\usepackage{array}
\usepackage{tabularx}
\title{Active Surface with Passive Omni-Directional Adaptation of Soft Polyhedral Fingers for \\In-Hand Manipulation}
\author{
  Sen Li\\
  School of Design\\
  Southern University of Science and Technology\\
  Shenzhen, China 518055\\
  \And
  Fang Wan$^*$\\
  School of Design\\
  Southern University of Science and Technology\\
  Shenzhen, China 518055\\
  \texttt{wanf@sustech.edu.cn}\\
  \And
  Chaoyang Song\thanks{Corresponding Author.}\\
  Department of Mechanical and Energy Engineering\\
  Southern University of Science and Technology\\
  Shenzhen, China 518055\\
  \texttt{songcy@ieee.org}\\
}
\begin{document}
\maketitle
\begin{abstract}

    Track systems effectively distribute loads, augmenting traction and maneuverability on unstable terrains, leveraging their expansive contact areas. This tracked locomotion capability also aids in hand manipulation of not only regular objects but also irregular objects. In this study, we present the design of a soft robotic finger with an active surface on an omni-adaptive network structure, which can be easily installed on existing grippers and achieve stability and dexterity for in-hand manipulation. The system's active surfaces initially transfer the object from the fingertip segment with less compliance to the middle segment of the finger with superior adaptability. Despite the omni-directional deformation of the finger, in-hand manipulation can still be executed with controlled active surfaces. We characterized the soft finger's stiffness distribution and simplified models to assess the feasibility of repositioning and reorienting a grasped object. A set of experiments on in-hand manipulation was performed with the proposed fingers, demonstrating the dexterity and robustness of the strategy.
    
\end{abstract}
\section{Introduction}
\label{sec:Intro}
    
    Manipulation and locomotion share morphological intelligence fostered by the evolutional adaptation to the complex environment, which can also be achieved in robotic systems through evolutionary reinforcement learning \cite{gupta2021embodied, Sun2023}. Land-based robotic locomotion takes the form of wheeled, legged, or tracked mobility \cite{Rubio2019}. Tracks are designed to distribute loads and enhance traction and maneuverability on loose surfaces, thanks to the larger contact areas, compared to wheeled and legged locomotions. Likewise, robotic in-hand manipulation tasks can also be categorized by how the fingers contact and drive the object to the target pose.

    Legged locomotion and precision grasp (with fixed contact at fingertip) for in-hand manipulation share the advantage of transversality and dexterity, respectively, and smaller footprint or contact area. Robotic grippers adopting such a strategy take the form of a multi-finger humanoid hand \cite{Abondance2020, openai2020, Dafle2014, McCann2017, UEDA2010} and require more dedicated perception and planning algorithms to achieve stability. A popular approach is to attach tactile sensors \cite{yuan2017gelsight, yan2021soft} to fingertips, further increasing the complexity and cost of the robotic hand. Like wheeled locomotion, a precision grasp with rolling contact aims to increase stability while maintaining dexterity through gripper design \cite{Yuan2020, Yuan2020b, Gomes2021}. 

    The counterpart to tracked mobility in manipulation uses active surfaces \cite{tincani2012velvet}. Grippers with active surfaces can drive the grasped object while maintaining continuous contact, providing higher stability and lower dexterity than the former two categories. Like the tracked mobility counterpart, most active surfaces encompass rigid or fully actuated fingers \cite{Ma2016, Kakogawa2016, Spiers2018}. Contrary to locomotion tasks where terrain usually varies at a larger spatial scale than the tracks, in-hand manipulation deals with objects smaller than or comparable to the finger size. The active surface over a rigid finger is inefficiently utilized, especially when the shape of an object is curved and only a tiny portion of the surface is in contact with the object. 

    Soft robotic hands \cite{shintake2018soft} made from compliant materials \cite{Thuruthel2019} or structures \cite{Wang2021} can adapt to the grasped objects, especially irregular and delicate ones, and achieve robustness at low cost. Recent work proposed an omni-adaptive soft network structure \cite{Wan2020,Yang2020}, which differs from fin-ray finger \cite{Shan2020,Xu2021} by providing a distinct twisting along the central axis. Their multi-finger grippers demonstrated robust power grasps. However, soft hands usually are challenging to control due to the infinite degrees of freedom, leading to a lack of dexterity. Recent works use compliant fingers with active surfaces \cite{Govindan2019,Cai2023} to increase dexterity in soft hands, which shows the potential to achieve simultaneous benefits of adaptability, stability, and dexterity. 

    This paper proposes a novel soft finger with both passive omni adaptation and active surface to achieve robustness and dexterity for in-hand manipulation. The modularized soft fingers can be easily mounted on a commercial rigid gripper, enabling it for compliant grasping and in-hand manipulation, as shown in Fig.1. Details of the design and experimental platform are described in Section \ref{sec:Methods}. The analysis of the grasping strategy is discussed in Section \ref{sec:Model}. Section \ref{sec:Results} demonstrates and evaluates the grasping and in-hand manipulation tasks on four types of objects. Finally, concluding remarks are included in Section \ref{sec:Discuss}.
    
\section{Gripper Design}
\label{sec:Methods}

\begin{figure}[!h]
	\includegraphics[width=0.5\columnwidth]{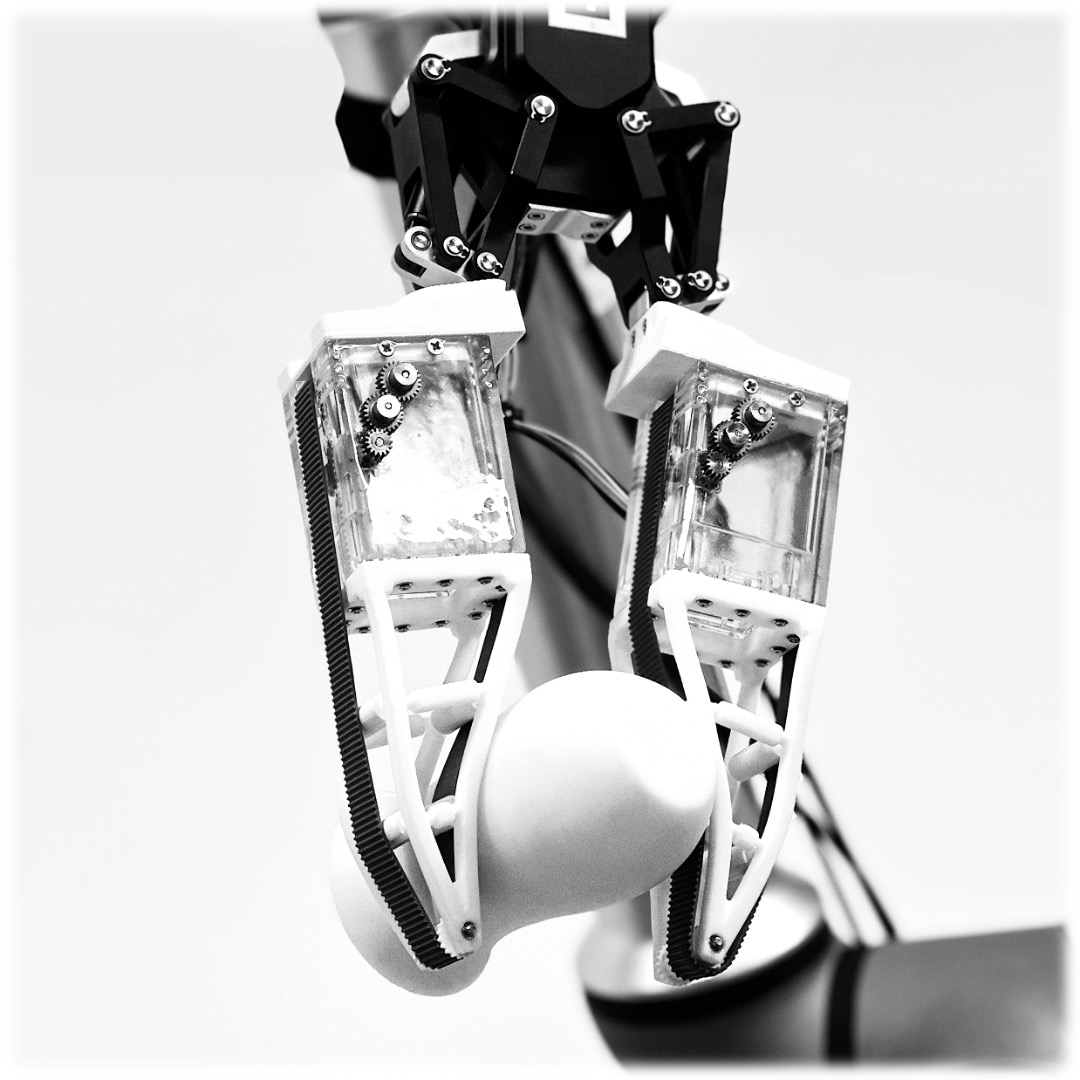}
	\centering
        \caption{The soft robotic finger with an active surface on an omni-adaptive network structure can be easily installed on existing grippers, achieving stable and dexterous in-hand manipulation.}
	\label{Fig1-prototype}
    \end{figure}

    To verify the hypothesis that it is possible to perform in-hand manipulation, an active surface design with passive omnidirectional adaptation was demonstrated by building the experimental platform described in this section. The prototype of the proposed in-hand manipulation technique, shown in Fig. \ref{Fig1-prototype}, have mainly three part: a DH-AG95 gripper, two passive soft polyhedral fingers, and two active surfaces.

\subsection{Passive Soft Polyhedral Finger Design}

\begin{figure*}[h]
        \centering
        \includegraphics[width=0.9\textwidth]{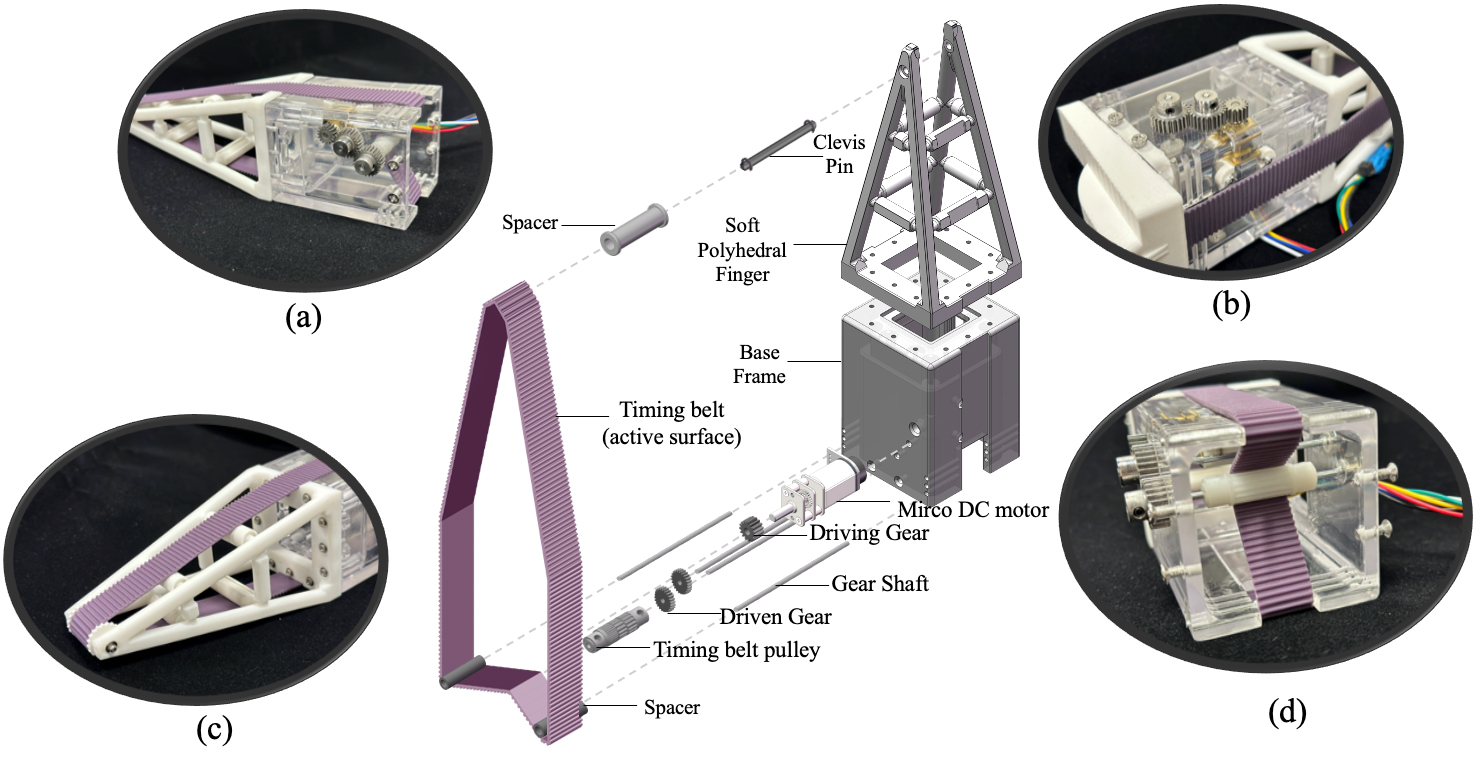}
        \caption{Assembly of mechanical components of the design: (a) Whole soft finger system, with active surface and passive omni-directional adaptation; (b) Physical power systems, encompassing motors and gearing mechanisms; (c) The active finger surface and the soft polyhedral finger; and (d) Active gear transmission mechanism in robotic systems.}
        \label{Fig2-Mech}
    \end{figure*}

    In this study, the passive soft polyhedral finger was designed using a universal approach. All edges of its polyhedron, featuring a pyramid shape at the top with two vertices, were transformed into soft-material beam structures. The beams at the top were replaced with clevis pins, through which passive rolling guidance was provided. Layers were then incorporated within the finger to establish a lattice. All ends of the mid-layer edges were redesigned into curved joints, allowing for a reduction in inferences during deformation while ensuring ample structural support. Fig. \ref{Fig2-Mech} illustrated the primary design features. Main interaction surfaces intended for typical grasping and secondary interaction surfaces designed for spatial adaptability, such as 3D twisting, are characterized in this design. We employed a polyurethane elastomer with a three-component mix ratio of 1:1:0 to attain consistent and stable performance. The entire lattice was fabricated through vacuum molding, achieving a hardness of 90A. The soft polyhedral finger in Fig. \ref{Fig2-Mech}(c) was chosen as the experimental platform because it has excellent 3D adaptations and fulfills the passive omnidirectional adaptation requirements. The soft polyhedral finger weighs 42 g, stands at a height of 125 mm, and has a base width of 54 mm.

    We mounted the finger on the base frame, which was 3D-printed from photosensitive resin and weighed 85 g. It also provides installation spaces for components such as the timing belt and the motor drive system.

\subsection{Active Surface Design}

    An active surface mechanism is incorporated in each soft polyhedral finger. Each active surface mechanism consists of four parts: a timing belt part, a timing belt pulley, a driving gear part, and a driven part.

    The timing belt, as depicted in Fig. \ref{Fig2-Mech}(c), functions analogously to a tank's track, influencing the motion of the grasped object. Each timing belt is constructed using polyurethane elastomers (TPU) with a hardness of 90A. With a modest weight of 10.33 g, its optimal stiffness and damping characteristics allow it to effectively exert a high contact force on objects. The timing belt pulley is manufactured through 3D printing from a high-toughness resin and weighs 2.08 g. The meshing tooth profile has been designed as a 2GT type. The timing belt has a width of 16.5 mm and a thickness ranging from 1 mm to a maximum of 2 mm.

    Comprising mainly a Mirco DC motor and the timing belt pulley, the driving part plays a pivotal role in the system. Each driving gear weighs 6.24 g. The driven element enhances system functionality to facilitate the smooth roll motion of the timing belt. Each driving gear has a mass of 1.66 g.

\section{Model and Analysis}
\label{sec:Model}

    This section presents details of the stiffness distribution and dynamics model. The related parameters of the whole soft finger system are listed in Table \ref{parameters}.

    \begin{table}[h]
        \caption{Specification of Model for anyalysis} 
        \begin{center}
        \begin{tabular}{ | c || c |  c|}
        \hline
        \textbf{Character}& \textbf{Definition}  \\
        \hline
        \hline
            $\theta_i$ & Distinct propulsion angles of the rig \\ 
                \hline
                $h_{i}$ & Height of position $i$ (m)\\ 
             \hline 
            ${^b}v$ &  Velocity of the grasped object (m/s)\\ 
             \hline      
            $v_{R}$ &  Velocity of the right timing belt (m/s)\\ 
             \hline 
            $v_{L}$ &  Velocity of the left timing belt (m/s)\\ 
                 \hline 
            $F_{N}$ &  Normal contact Force between belt and object (N)\\ 
                 \hline 
            $F_{S}$ & Shear contact Force between belt and object (N)\\ 
                 \hline 
            $\alpha$ &   Rotation angle (rad)\\
                 \hline      
            $r$ &  Radius of a cylinder or sphere (mm)\\ 
                 \hline 
                     $S_R$ & Displacement of the right timing belt (m)\\ 
                 \hline 
                     $S_L$ & Displacement of the left timing belt (m)\\  
                 \hline 
            $\mu_{bo}$ &  Friction coefficient of the belt\\ 
                 \hline 
            $m_c$ &  Grasped object mass (kg)\\  
                     \hline 
                $I$ &  Grasped object mass moment of inertia  \\    
         \hline      
        \end{tabular}
        \label{parameters}
        \end{center}
    \end{table}

\subsection{Reposition in Simplified Model}

    To pull an object closer to the finger base for a firmer grasp or to push it towards the fingertips to expose a specific part for usage or grasp-release operations, the timing belts can be moved synchronously in the same direction, facilitating the repositioning of the object between the fingers.

    \begin{figure}[h]
        \centering
        \includegraphics[width=0.5\columnwidth]{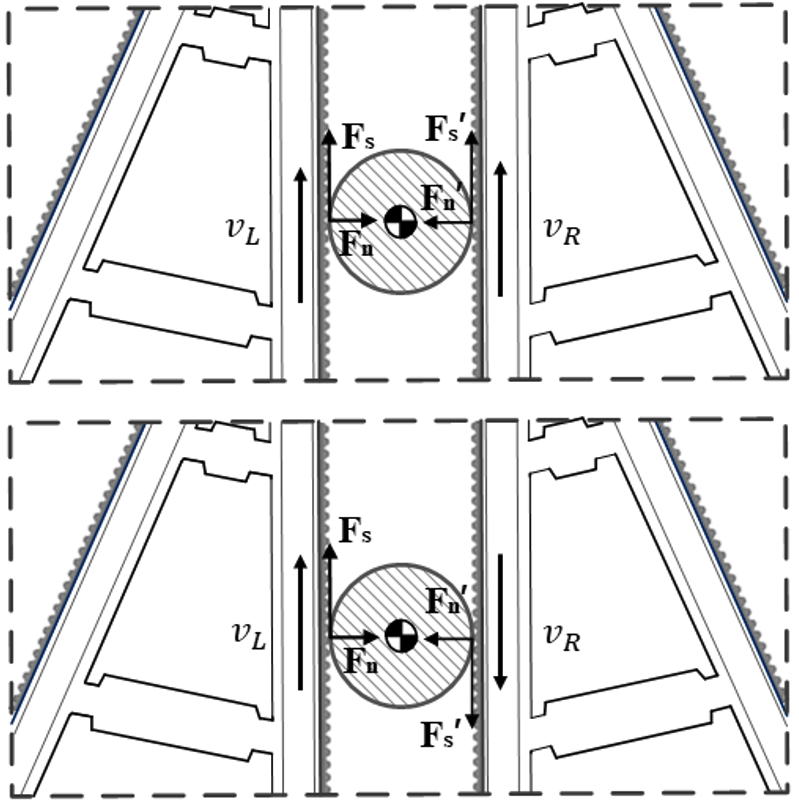}
        \caption{Contact model between the grasped object and the belt to reposition (Top) and reorient (Bottom) target object.}
        \label{contactforce}
    \end{figure}

    Fig. \ref{contactforce} shows the object held by two fingers. $b$ donates the body frame of the grasped object. The hypotheses for this simplified model are as follows: a) friction is present between the object and the dual belts, b) the friction coefficient of the belts is a known constant $\mu_{bo}$, c) air resistance, is disregarded in this model, and d) the upward motion of the belt is conventionally defined as the positive direction. The dynamics Eq. of this system is:
    \begin{equation}
        m_c {^b}\ddot{x}=\left\{
        \begin{array}{rrcl}
        F_s+F_{s'}-m_c g  & v_L \textgreater 0 \land  v_R \textgreater 0 \land v_L,v_R \geq {^b}v\\
        -F_s-F_{s'}+m_c g & v_L \textless 0 \land  v_R \textless 0 \\
        \end{array} \right.
        \label{eqn1}
    \end{equation}

    The equilibrium between the dual frictional forces and gravitational force primarily determines the vertical motion of the object. When the belts accelerate upwards, and if the resulting frictional force surpasses the gravitational pull on the object, the object will accelerate upwards in unison with the belts. Conversely, if the belts accelerate downward and the frictional force is insufficient to counteract the object's weight, the object will descend with the belts. In scenarios where the belts decelerate or come to a halt, the diminished frictional force might not be adequate to sustain the object's weight, leading it to slide downward unless otherwise counteracted.

    Assuming an object is grasped at the midpoint between the two fingers, $G_{max} = \frac{2\mu_{bo} \tau_m  }{L_c} $ is estimated to be the maximum weight of the grasped object. Here, $\mu_{bo}$ represents the friction coefficient between the timing belt and the object, $\tau_m = 3.4$ kg$\cdot$ cm is the maximum torque output of the motor, and $L_c$ is the distance between the motor's output shaft and the point of contact on the finger. Taking $\mu_{bo} =0.5 $, the maximum weight of the object that can be grasped is determined to be approximately 0.57 kg.

\subsection{Reorientation in Simplified Model}

    Owing to the differential velocities of the two belts, the frictional forces they exert induce a rotation about the object's axis. Treating the object as a rigid body undergoing rotation about a fixed axis, its moment of inertia, $I$, can be taken as a known value. The object's angular acceleration, $\ddot{\alpha}$, can be determined by the torque, $\tau$, and its moment of inertia. The torque arising from frictional forces can be characterized using the frictional force and the object's radius, $r$. Given the friction exerted by the belts on the object, the resulting torque can be expressed:
    \begin{equation}
        I\ddot{\alpha} =r \times (F_s - F_{s'})
    \end{equation}

    Thus, the relative velocities of the conveyor belts and their friction coefficients dictate the grasped object's rotational velocity and direction. Meanwhile, the grasped object's vertical motion is governed by the balance between its gravitational force and the combined frictional forces from both belts.

    When the two timing belts move at different speeds in opposing directions ($v_L\cdot v_R\geq 0$), differential contact forces are exerted between the object and the two belts: one upwards and the other downwards. In particular, consider a specific scenario where these contact forces constitute a force couple, a pair of forces acting on the grasped object. This force couple inherently tends to induce rotation in the object. Given this configuration, the principle behind the rotation of the grasped object becomes intuitively evident.

    The degree of rotation can also be calculated based on the displacement of the timing belts, as shown in Eq. \ref{rotation}, where $\eta = \frac{\pi}{180}$. 
    \begin{equation}
        \alpha=\left\{
        \begin{array}{rrcl}
        \frac{ r}{(S_L-S_R)}\cdot \eta & v_L\cdot v_R \textgreater 0\\
        \frac{ r}{(S_L+S_R)}\cdot \eta\qquad & else\\
        \end{array} \right.
        \label{rotation}
    \end{equation}

    \begin{figure}[t]
        \centering
        \includegraphics[width=0.6\columnwidth]{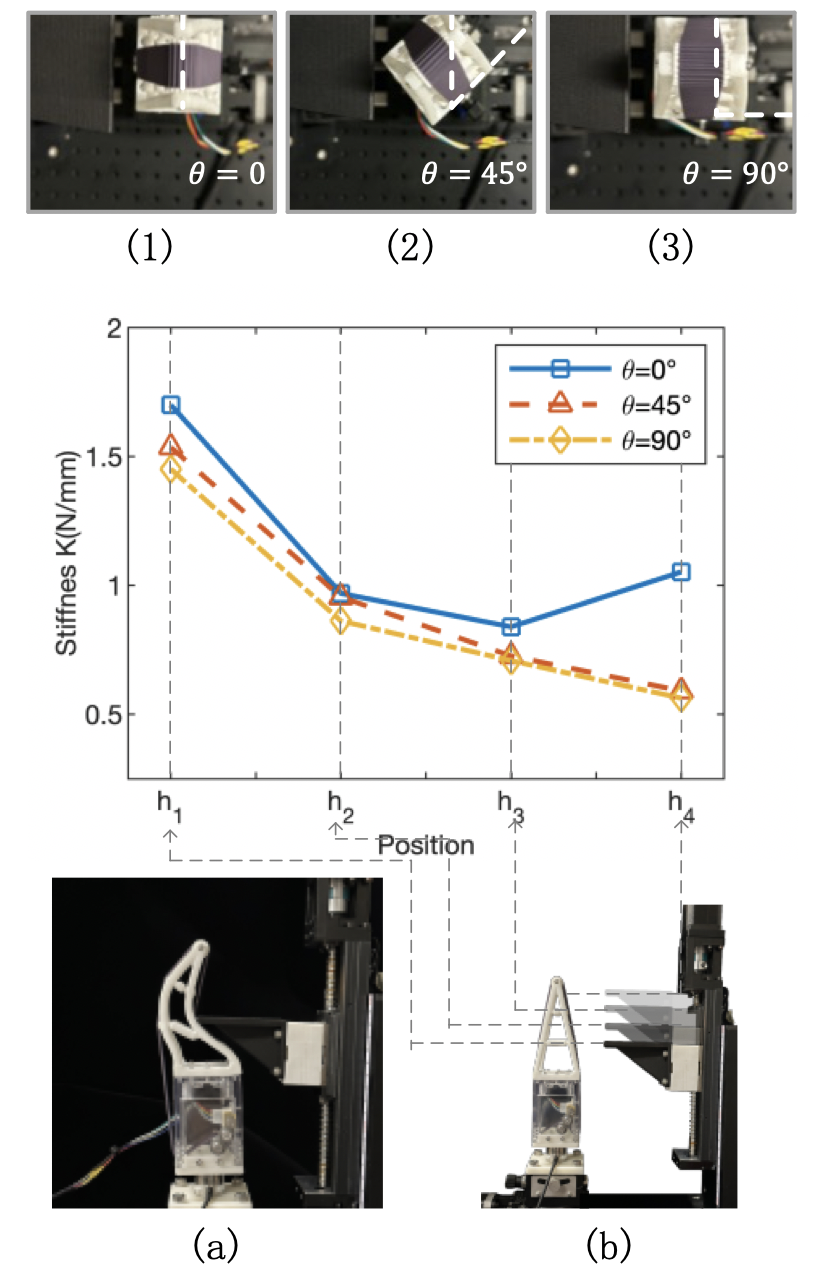}	
        \caption{The stiffness distribution of the soft polyhedral finger was measured using the test rig. (Top) Three distinct propulsion angles, (Middle) stiffness distribution of soft polyhedral finger with an active surface, (Bottom left) an experiment trial at $\alpha = 0^{\circ}, h= h_1$, and (Bottom right) experiment setup for measuring stiffness.}
        \label{Stiffness}
    \end{figure} 

\subsection{Passive Finger Compliance: Stiffness Distribution}

    A series of unidirectional compression tests were conducted to estimate the stiffness distribution of the soft polyhedral finger, characterized by the relationship between force and displacement. As illustrated in Fig. \ref{Stiffness} b, the soft finger is mounted atop a custom test rig, equipped with a high-performance force sensor (ATI Nano25), featuring two motorized linear motions and two manual rotary motions. The probe compressed the soft polyhedral finger horizontally at a rate of 5 mm/s with a predefined depth of 15 mm. Tests were conducted at three distinct propulsion angles.

    Experiments were carried out at three distinct propulsion angles: $\theta=0^{\circ}$, $\theta=45^{\circ}$, and $\theta=90^{\circ}$, shown as Fig. \ref{Stiffness}. The measurement results exhibited consistent trends where a decrease in stiffness distribution suggested that the edges of the soft finger possess moderate adaptability, showcasing adaptiveness in three-dimensional space. For all experiments, Fig. \ref{Stiffness} depicts the decline of stiffness distribution along the z-axis from $h_1$ to $h_4$ for angles $\theta=45^{\circ}$, and $\theta=90^{\circ}$. However, at an angle $\theta=0^{\circ}$, the stiffness decreases from 1.699 N/mm at $h_1$ to 0.838 N/mm at $h_3$, with a slight increase to 1.051 N/mm at $h_4$. This unique stiffness distribution contrasts with the fin-ray effect fingers, where the stiffness at the fingertip drops to approximately 25\% of the stiffness near the base.

    \begin{figure*}[t]
        \centering
        \includegraphics[width=0.8\textwidth]{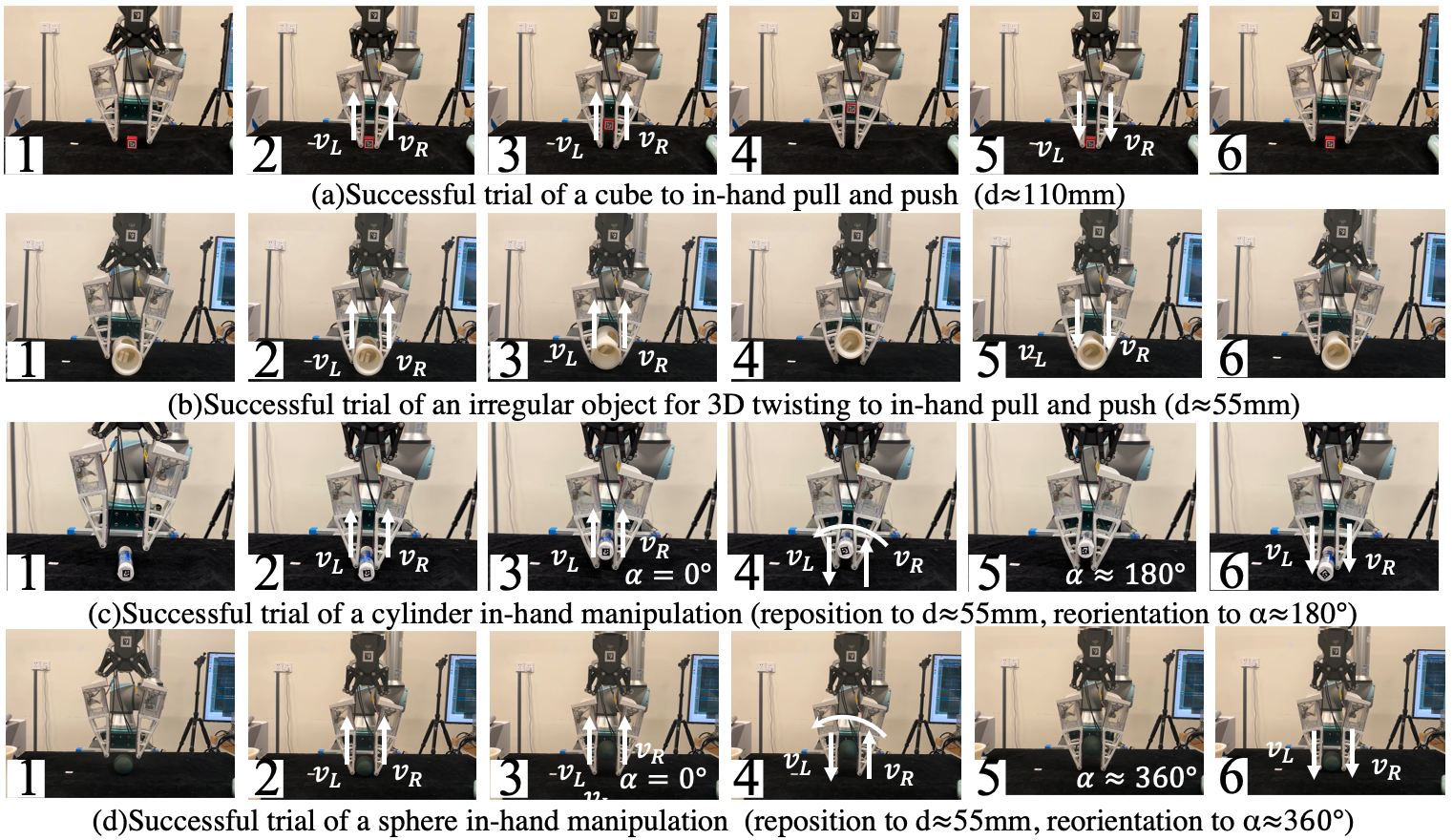}	
        \caption{Back-to-back perching experiment (left to right, top to bottom). (a) and (b) experimental initial condition: displacement was set to 110 mm and 5 5mm, respectively. (c) and (d) experimental initial condition: displacement was set to 55 mm, rotation angle was set to 180$^{\circ}$ and 360$^{\circ}$, respectively. }
        \label{experiment}
    \end{figure*} 

\section{Results and Discussions}
\label{sec:Results}

    Experiments were performed on four objects, including a cube, an irregular 3D-printed vase, a cylinder, and a sphere to confirm that the proposed strategy can enable in-hand manipulation. Detailed information regarding the items grasped during the experiments is summarized in Table \ref{Products}.

    \begin{table*}[h!]
        \caption{Objects used for the in-hand manipulation tests.}
        \centering
        \begin{tabular}{|c|c|c|c|c|}
            \hline
            Property/Shape & Cube & Irregular Object  & Cylinder & Sphere \\
            \hline
            \hline
            Picture{} & \includegraphics[width=3cm]{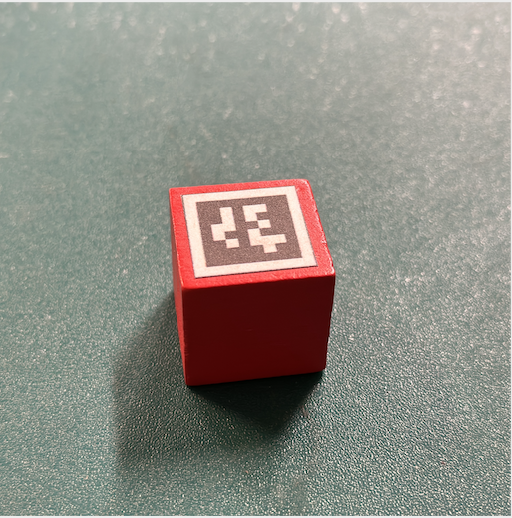} & \includegraphics[width=3cm]{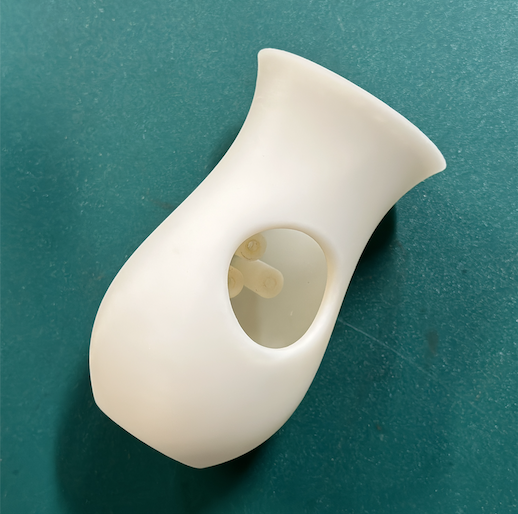}   & \includegraphics[width=3cm]{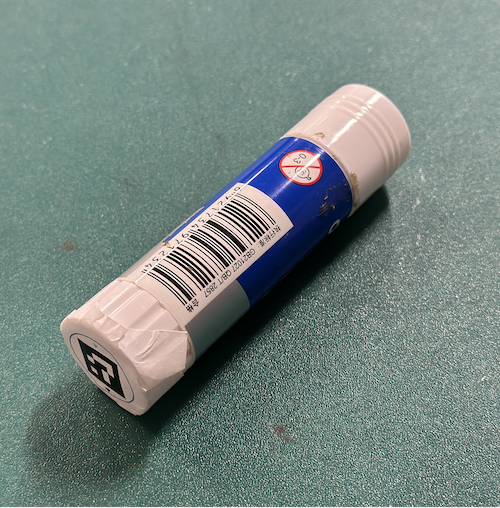}   & \includegraphics[width=3cm]{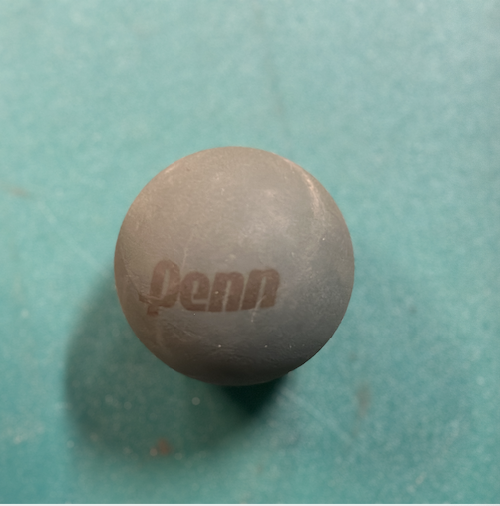}  \\
            \hline
            Weight(g) & 8.66  & 132.04  & 31.74   & 42.15 \\
            \hline
            Length(mm) & 25.00 & 120.00  &112.70   & 55.00 \\
            \hline
        \end{tabular}
        \label{Products}
    \end{table*}
    
\subsection{Repositioning}

    As shown in Fig. $\ref{experiment}$(a) and $\ref{experiment}$(b), the repositioning strategy was successfully applied to a cube and an irregular 3D-printed vase respectively. In the successive snapshots of a successful repositioning, the contact occurs, and the object interacts with the belt as it is pressurized. From the second to the fourth snapshot, as the entire object was lifted up, the kinetic energy of the belt was converted into the gravitational potential energy of the grasped object. In the fourth panel, the pull-in phase terminates, and the push-out phase begins. The object falls under the controlled belt in the fifth and sixth panels. We further tested the robustness of repositioning by conducting each experiment 20 times. The cube was successfully repositioned to the middle segment 20 times out of 20 trials with an angular velocity of the belt of approximately 60 rad/s. The irregular vase was successfully grasped and repositioned in 14/20 trials with an angular velocity of the belt of approximately 60 rad/s.

    \begin{figure}[h!]
        \centering
        \includegraphics[width=0.6\columnwidth]{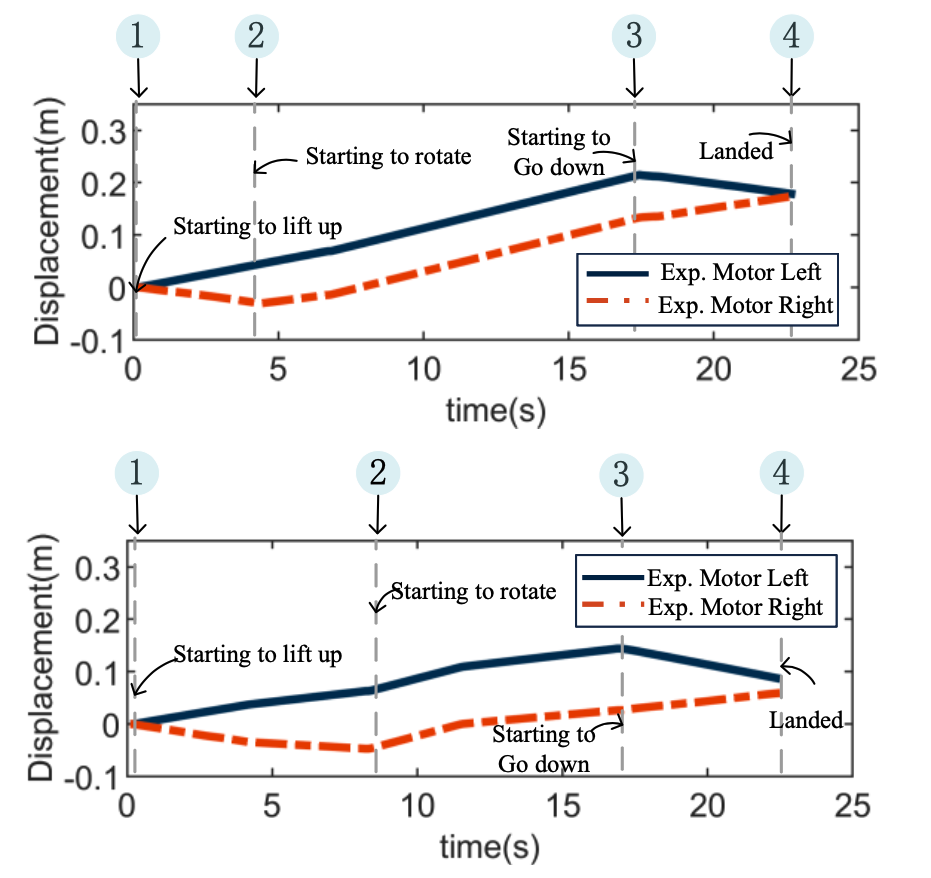}	
        \caption{Experimental displacement data were obtained from successful trials involving (Top) a cylinder and (Bottom) a sphere. The key timing is as follows: contact and start to lift up at \ding{172}, orientation begins at \ding{173}, downward movement happens during \ding{174}-\ding{175}, and stable placement at \ding{175}.}
        \label{displacement}
    \end{figure}    

\subsection{Repositioning and Reorientation}

    Repositioning and Reorientation were tested on a cylinder and a sphere. Figs. $\ref{experiment}$(c)\&(d) show successive snapshots of successful in-hand manipulations of a cylinder and sphere, respectively. In the second snapshot of Figs. $\ref{experiment}$(c)\&(d), the contact begins, and the object interacts with the belt as it is pressurized. From the second to the third snapshot, as the entire object was lifted up to the middle of the finger, the kinetic energy of the belt was converted into the gravitational potential energy of the grasped object. In the fourth snapshot, the repositioning phase terminates, and the reorientation phase begins. In the fifth snapshot, the orientation phase terminates, and the push-out phase of orientation begins. The object falls under the controlled belt in the fifth and sixth snapshots, while the gravitational potential energy is converted to kinetic energy. The displacement data of the two motors in Fig. \ref{displacement} shows the movement differences. 
    
    As shown in Fig. \ref{experiment}(c), the cylinder was successfully reorientated in 16/20 trials with an angular velocity of the belt of approximately 60 rad/s. As shown in Fig. \ref{experiment}(d), the sphere was successfully reorientated in 20/20 trials with an angular velocity of the belt of approximately 60 rad/s. The experimental results verify that the proposed in-hand manipulation strategy works effectively while maintaining a stable grasp. Fig. 
    \ref{displacement} indicates that the trend of the displacement at a cylinder and the sphere is consistent. The grasped object rotated after reaching the middle of the soft polyhedral finger and started to rotate under the effect of gravity and the contact force. 

\section{Conclusion and Future Works}
\label{sec:Discuss}

    This paper presents the design of a soft finger aimed for stability and dexterity for in-hand manipulation. This is achieved by two key design strategies: the soft polyhedral network structure with omni-adaptability, and the active surface covering the soft structure offering extra dexterity. Two proposed fingers were used to augment a commercial rigid two-finger gripper and enable it to perform in-hand manipulation. With active surfaces, the grasped object can be repositioned and reoriented without the need for environmental assistance or reliance on push-and-grasp techniques. The omni-adaptability extends previous works on active surfaces by enhancing grasping stability and flexibility. Experiments on the in-hand manipulation tasks show that the finger design works well for objects with smooth surfaces and irregularly shaped items like vases.

    There were some trials in which the object failed to be in hand. The main reason may be that the beam of the soft finger was too convex when the contact happened, causing one side of the object to move faster than expected. This study focused on emulating the adaptive behavior of track systems for irregularly shaped objects without directly imitating its power system. Some qualitative conclusions are that, as the contact area and force increase, the belt's width and the groove's depth must follow to ensure that. However, such a design may not be realistic for belts to handle large loads because of the excessive motor torque required. As a result, the relationship between the corresponding parameters related to the scale factor of the belt should be further studied. 
    
\bibliographystyle{unsrt}
\bibliography{References}  
\end{document}